\documentclass[conference]{IEEEtran}
\IEEEoverridecommandlockouts
\usepackage{times}
\usepackage[hyphens]{url}

\usepackage[numbers]{natbib}
\usepackage{multicol}
\usepackage[bookmarks=true]{hyperref}
\usepackage{graphicx}
\usepackage{booktabs}

\begin{document}

\title{Moderating Group Conversation Dynamics\\with Social Robots}


\author{Lucrezia~Grassi$^1$, Carmine~Tommaso~Recchiuto$^1$,~Antonio~Sgorbissa$^1$
\thanks{$^1$All authors are with the University of Genoa, DIBRIS, Via All'Opera Pia 13, 16145, Genoa, Italy.}
\thanks{Corresponding author's email: \texttt{\href{mailto:lucrezia.grassi@edu.unige.it}{lucrezia.grassi@edu.unige.it}}}
\vspace*{-8mm}
}



%

\maketitle

\begin{abstract}
This research investigates the impact of social robot participation in group conversations and assesses the effectiveness of various addressing policies. The study involved 300 participants, divided into groups of four, interacting with a humanoid robot serving as the moderator. The robot utilized conversation data to determine the most appropriate speaker to address. The findings indicate that the robot's addressing policy significantly influenced conversation dynamics, resulting in more balanced attention to each participant and a reduction in subgroup formation.
\end{abstract}

\IEEEpeerreviewmaketitle

\section{Introduction}
Social robotics focuses on creating and applying robots designed to engage with humans in social environments. The use of social robots in group interactions is attracting increasing interest due to the growing need for these robots to engage with multiple users simultaneously. This includes scenarios such as robots participating in discussions with several individuals \cite{gillet2022}, facilitating social interactions \cite{mataric2017}, and engaging in social games \cite{pereira2014}.

Robots capable of engaging with multiple people can enhance user experiences by making interactions more compelling and natural. This can lead to higher user satisfaction and improved accessibility and inclusivity, as highlighted by \cite{tuncer2022}. However, at present, few robots can engage with multiple users simultaneously. This is due to the many factors that need to be considered in these situations. A system described in \cite{yumak2014} examines the tracking and fusion aspects of multi-party interactions but only monitors user entry and exit and can accurately identify only two users. A spoken dialogue system like the one in \cite{pappu2013} can identify multiple users using data from a Kinect sensor, but its conversational abilities are limited, and it struggles with long, natural conversations involving multiple parties. Additionally, it can only engage one person at a time. Similar limitations are observed in the work of \cite{moujahid2022}, which aims to develop a multi-user engagement policy for managing turn-taking using the robot's gaze, head movements, and verbal communication. 

Whenever the robot interacts with multiple people, the “many minds problem'' arises \cite{cooney2020}. As the number of participants increases, basic conversational mechanisms such as turn-taking, speaking time, and listener feedback become more complex. Although turn-taking is a fundamental aspect of communication, researchers continue to study how speakers signal the end of their turn and indicate who they are addressing, as well as how listeners recognize when it is their turn to speak. This is accomplished through behaviors such as gaze, head orientation, and intonation, which have been studied by various scholars, including \cite{deruiter2006, duncan1972, riest2015}. Speakers typically use gaze to select the next speaker, known as the “addressee'' \cite{clark1982, bohus2010, vertegaal2001}. The selection of the addressee is a crucial issue in group conversations, requiring speakers to make quick decisions that consider the potential reactions of other participants.

\begin{figure}[t]
    \centering
    \includegraphics[width=0.98\linewidth]{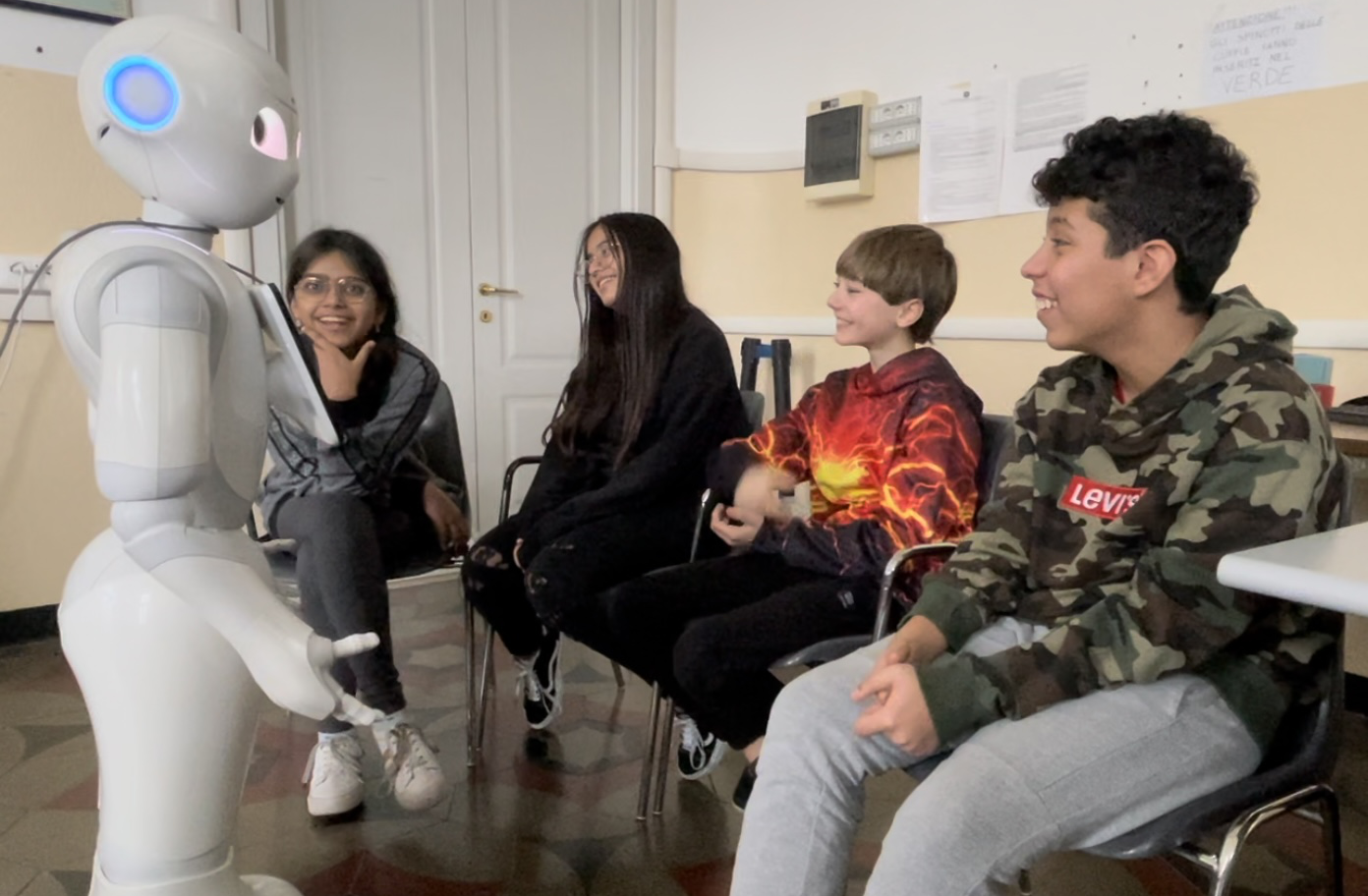}
    \caption{Multi-party interaction between the humanoid robot Pepper and a group of four students of the “Parini Merello'' middle school in Genoa, during an experiment performed for this study.}
    \label{fig:interaction}
\end{figure}

Furthermore, the robot needs to comprehend the participation levels of the users it is interacting with. Dominance, a key concept in social interactions, plays a significant role in this understanding and has been extensively studied in social psychology \cite{ellyson1985, burgoon2006}. Dominance can pertain to an individual's traits or their hierarchical status within a group. Indicators of dominance are categorized into vocalic and kinesic types. Vocalic indicators include factors such as speaking time, word count, and speech loudness (or energy) \cite{schmid2002}. Speaking activity, especially the duration of speech, is a strong predictor of dominance. Kinesic indicators involve body movement, posture, facial expressions, and eye gaze \cite{dunbar2014}. Dominant individuals generally exhibit more movement and a broader range of motion than non-dominant individuals, and their gestures during speech are positively associated with dominance \cite{burgoon2006}.


This paper presents a study that aims to investigate and regulate the dynamics of group conversations involving a social robot. In such interactions, participants may have varying roles and personalities, leading some individuals to dominate the conversation while others may feel excluded. Ensuring balanced participation is essential to make everyone feel included and engaged.

To manage the dynamics of the conversation, particularly when it is crucial for all participants to feel like part of a cohesive group, we implemented four different control policies that leverage the concepts of dominance and communities. To validate these policies, we tested them against a baseline policy. We conducted 75 experiments involving a total of 300 participants, where a humanoid robot engaged in conversations with groups of four individuals. Figure \ref{fig:interaction} depicts the robot interacting with the participants in one of the experiments. Throughout these experiments, the robot collected quantitative data to analyze participation levels, identify distinct subgroups (i.e., communities) of participants, and evaluate the overall conversation dynamics. This information allows the robot to act as a moderator, promoting active participation in the conversation.

The article is structured as follows. Section \ref{sec:methodology} provides an overview of the system architecture and the developed control policies. Section \ref{sec:exp-results} describes the experimental setup and discusses the findings. Finally, Section \ref{sec:conclusion} presents the conclusions.

\section{Methodology}
\label{sec:methodology}
\subsection{System Architecture}
The robot's ability to converse autonomously with multiple people is enabled by CAIR (Cloud Artificial Intelligence and Robotics), a cloud software architecture specifically designed for autonomous conversation \cite{recchiuto2020, grassiAIRO2021}. This system relies on a framework for knowledge representation, utilizing an ontology implemented in OWL2. Conversation topics and related sentence pieces are dynamically composed at runtime using the hierarchical structure of the knowledge base.

\begin{figure}
    \centering
    \includegraphics[width=\linewidth]{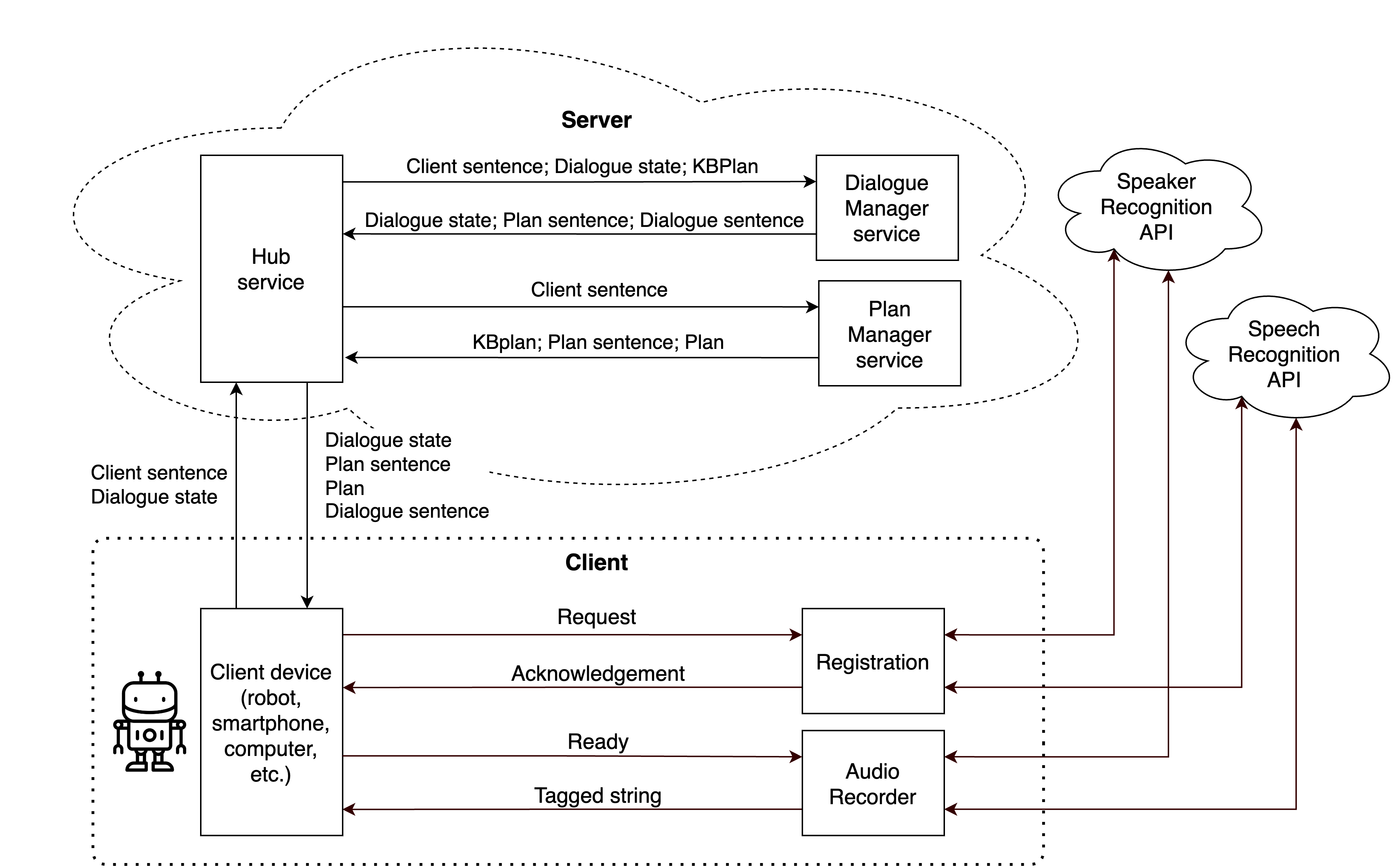}
    \caption{CAIR system architecture}
    \label{fig:architecture}
\end{figure}

The CAIR server comprises a set of web services, as illustrated in Figure \ref{fig:architecture}. The Dialogue Manager service manages the conversation and identifies the user's intent to discuss specific topics, while the Plan Manager service interprets the user's intent to direct the robot to perform particular actions. To generate appropriate responses and plans, the server uses an Ontology that includes all the topics, keywords, sentences, and plans used during interactions with users \cite{recchiuto2, grassiSocRob2022}. An additional service called Hub handles all incoming requests by forwarding them to the Dialogue Manager and Plan Manager services. Information exchange between the CAIR client and server is facilitated through the dialogue state, which tracks the conversation's history and includes discussed topics, preferred topics based on user input, and previously spoken sentences to avoid repetition.

To enable multi-party interaction, the original architecture described in \cite{grassiAIRO2021} has been enhanced with two new services: the Registration service and the Audio Recorder service. The Registration service is activated when a new user initiates registration, creating a new profile ID linked to the user's voice. The Audio Recorder service starts recording audio when the Root Mean Square (RMS) of the noise exceeds a specific threshold, sending the audio to the Speech Recognition API for transcription and to the Speaker Recognition API to identify the speaker's profile ID.

The dialogue state has been expanded to support multi-party interaction, incorporating data related to the speakers along with dialogue statistics. These include a matrix tracking the number of times one speaker talked after another in successive turns, the total number of turns for each user, the average topic distance between speakers, the a priori probability that a speaker will talk, and a moving window that tracks the information related to speakers' turns. The moving window maintains the most recent conversation turns within a duration equal to $M$. For each turn, it records the speaker's ID, speaking time, and word count. When the total speaking time in the moving window exceeds $M$ minutes, the earliest turn is removed, and the latest one is added (FIFO queue).

\subsection{Control Policies}
The information contained in the moving window has been used to develop two policies aimed at controlling different aspects of group dynamics.

\subsubsection{Balancing Policy}
The Balancing policy exploits the data from the moving window to determine which speaker to address. Its goal is to identify and engage with the user who is least active in the conversation (i.e., the submissive user). Conversation participation is quantified using a metric $D_i$, which considers both speaking time and word count, as these are the most significant indicators of dominance \cite{schmid2002}. To calculate $D_i$ for each speaker $S_i$, we measure the percentage of their speaking time ($T_i$) and word count ($W_i$) within the moving window. The metric $D_i$ is then computed as:
\begin{equation}
   D_i = \gamma_1 T_i + \gamma_2 W_i,
\end{equation}
where $\gamma_1$ and $\gamma_2$ are weights representing the relative importance of speaking time and word count in determining dominance. The speaker addressed by the Balancing policy is $S_m$, where:

\begin{equation}
m = \arg \min_i(D_i).
\end{equation}

There are two versions of this policy: the “hard" version (BH) and the “soft" version (BS). In the hard version, if a user other than the intended one responds, the robot ignores the response and repeats the question to the intended user. In contrast, the soft version accepts responses from any user, replies accordingly, and then readdresses the intended user with the original question.

\subsubsection{Community Policy}

The Community policy is based on the hypothesis that it is possible to identify sub-groups (i.e., communities) among participants in a conversation. To identify these communities, we use the probability that one speaker talks after another, assuming that members of the same community tend to speak consecutively. This probability is derived from the matrix in the dialogue state, which tracks the frequency with which each speaker follows another. This data is represented in a matrix and then converted into an undirected graph, where nodes represent speakers and edge weights represent probabilities. The Louvain algorithm is applied to this weighted graph to find the optimal partitioning of nodes into communities \cite{que}. Once the best partition is identified, the policy uses this information to address a random speaker from a different community at each turn. The goal is to maintain a single cohesive conversation and prevent speakers from dividing into sub-groups. Similar to the Balancing policy, both a hard (CH) and a soft (CS) version have been developed for the Community policy.

Finally, it is important to note that the policy to be used by the system can be selected before the start of the interaction.

\section{Experiments and Results}
\label{sec:exp-results}
The experiments were conducted at the “Parini Merello" middle school in Genoa, where we brought the humanoid robot Pepper, equipped with the CAIR cloud client. Approval was obtained from the University of Genoa's ethical committee, and consent forms were signed by the parents of all participating first and second-grade students. A total of 300 students participated, divided into 75 groups of four. Each group took part in an experiment where the robot used one of the five developed policies, resulting in five types of experiments with 15 groups per experiment.

To establish a baseline, 15 groups interacted with the robot using a Neutral policy (N). The remaining groups were equally split among the other four policies. Participants registered to CAIR using their voices and then interacted freely with the robot for 15 minutes. They were unaware of the specific policy in use and could respond at their discretion. This approach ensured spontaneous and natural interactions. Each experiment required approximately 5 minutes for registration and 15 minutes for interaction with the robot. To ensure the privacy of the participants, all their personal information, including name, gender, and biometric data related to their voice, was promptly deleted after each experiment.

For the control groups (N), no specific results are expected, as this baseline represents the robot not attempting to influence the conversation dynamics. In contrast, for the first and second experimental groups (BH and BS), we anticipate that the robot will successfully “balance" participation, ensuring all participants are included equally in terms of speaking time and word count. For these experiments, we set the weights $\gamma_1 = \gamma_2 = 0.5$. The third and fourth experimental groups (CH and CS) introduce the concept of community, aiming to demonstrate the robot's ability to unify any emerging sub-groups among the participants.

\begin{figure}[t]
    \centering
    \includegraphics[width=0.45\textwidth]{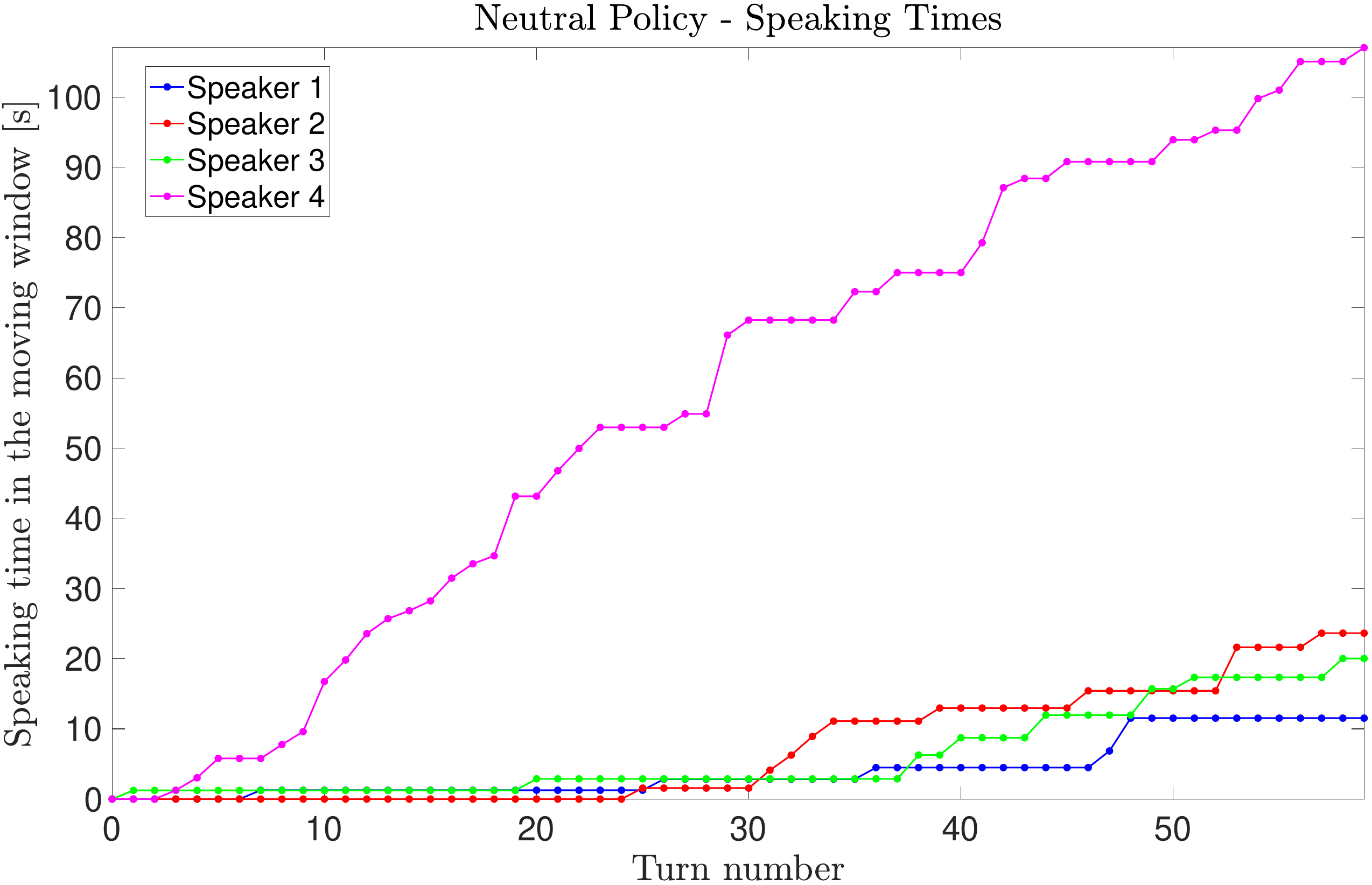}
    
    \vspace{2mm}
    
    \includegraphics[width=0.45\textwidth]{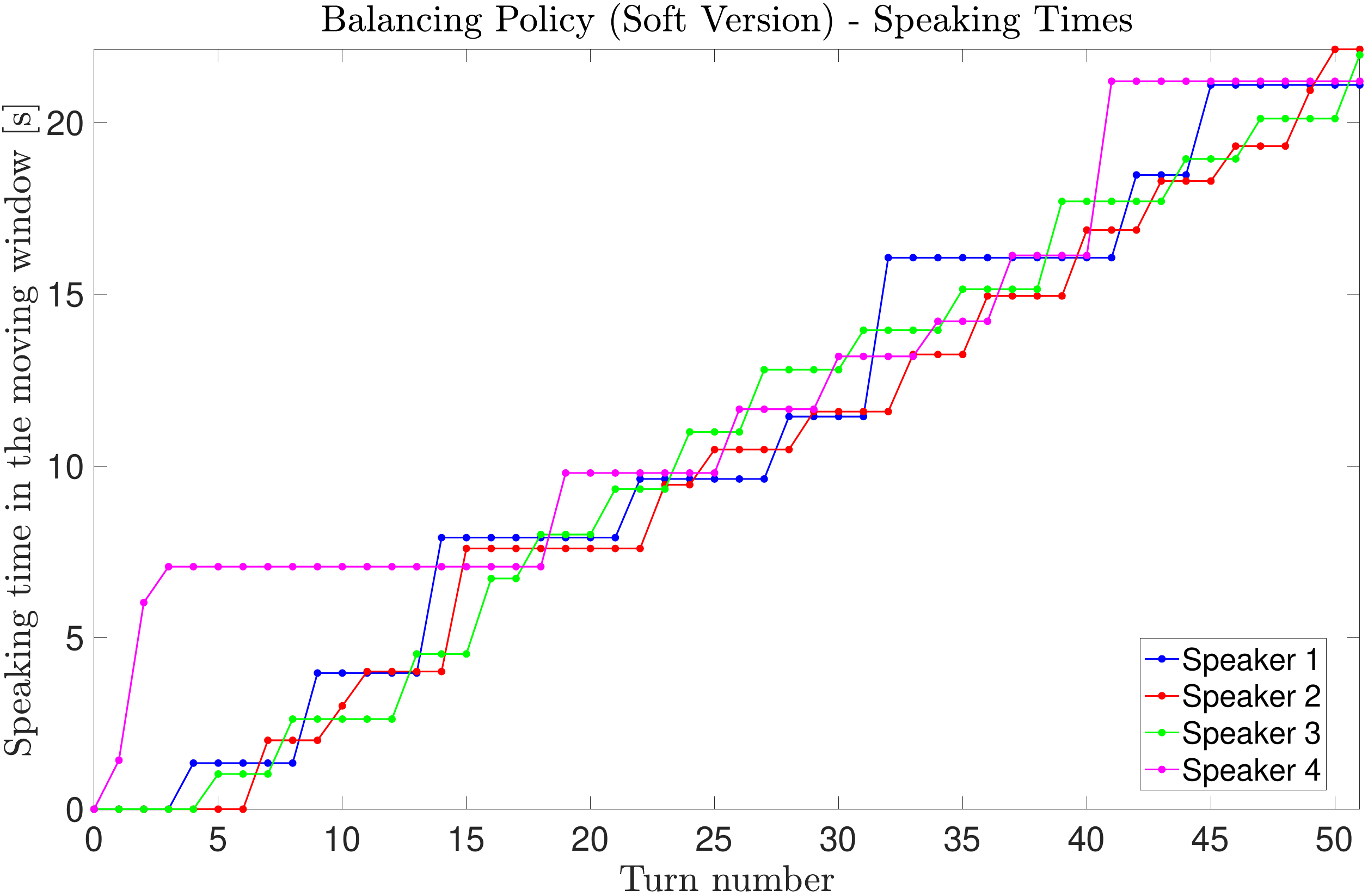}
    \caption{Comparison of the speaking times of the participants when interacting with the robot using the Neutral policy and the Balancing policy in its soft version.}
    \label{fig:times}
\end{figure}

Figure \ref{fig:times} illustrates the speaking times of four participants from two separate experiments using the N policy and the BS policy, respectively. The x-axis represents the turn number, while the y-axis shows the speaking time (in seconds) within the moving window. These results exemplify how, under the N policy, a dominant speaker often emerges, with the difference between the highest and lowest speaking times widening. This occurs because the robot does not direct the conversation, allowing extroverted participants to dominate. Conversely, when the robot uses the BS policy, even its soft version, the speaking times of participants remain more balanced.

It is important to note that the number of turns varies across experiments. This variation is primarily because, although all experiments have a duration of $15$ minutes, participants may speak for different amounts of time or remain silent due to initial shyness or uncertainty about what to say. Typically, the number of turns observed in the experiments ranged from about $40$ to $60$.

\begin{figure}[t]
    \centering
    \includegraphics[width=0.45\textwidth]{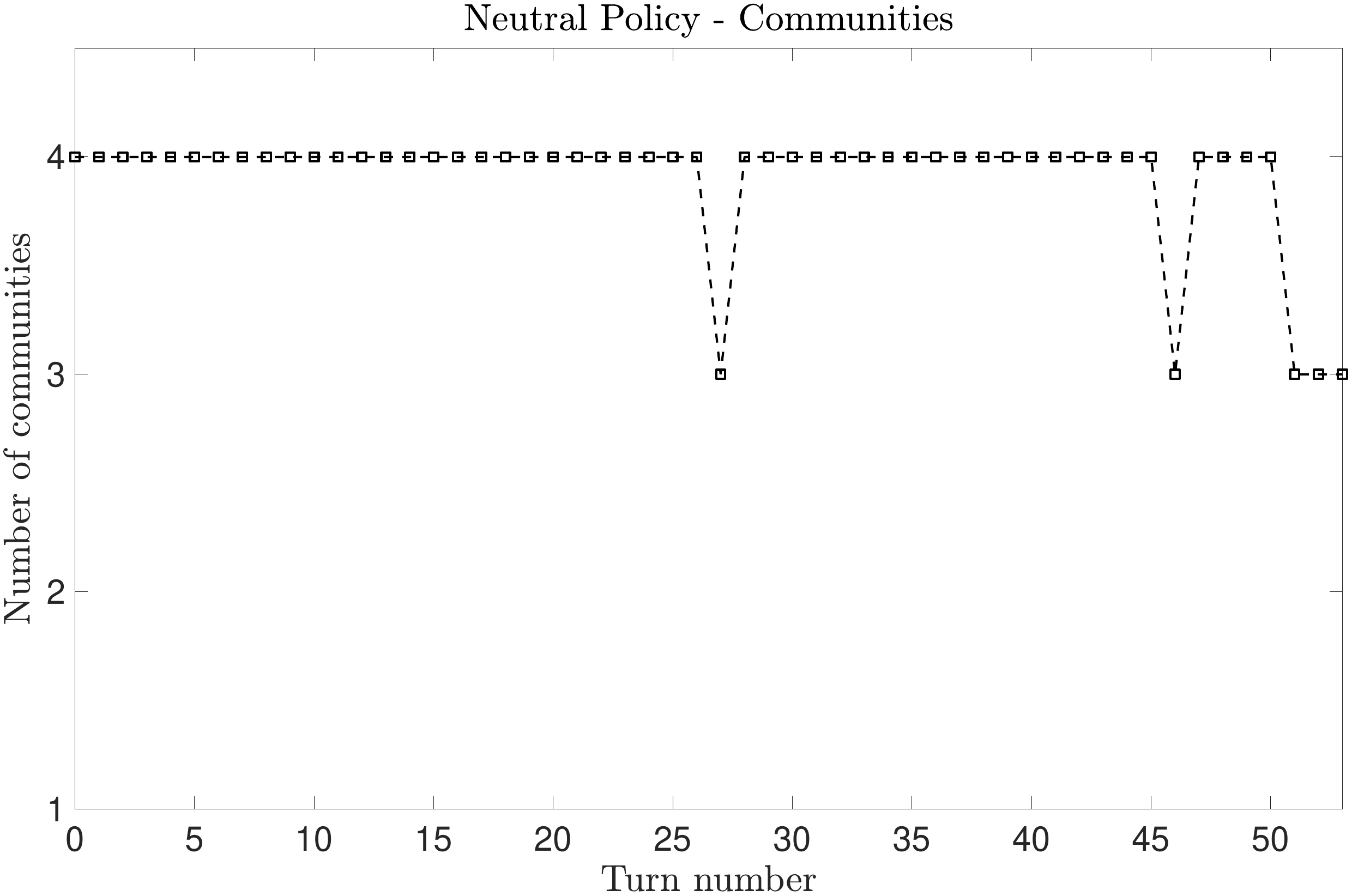}
    
    \vspace{2mm}
    
    \includegraphics[width=0.45\textwidth]{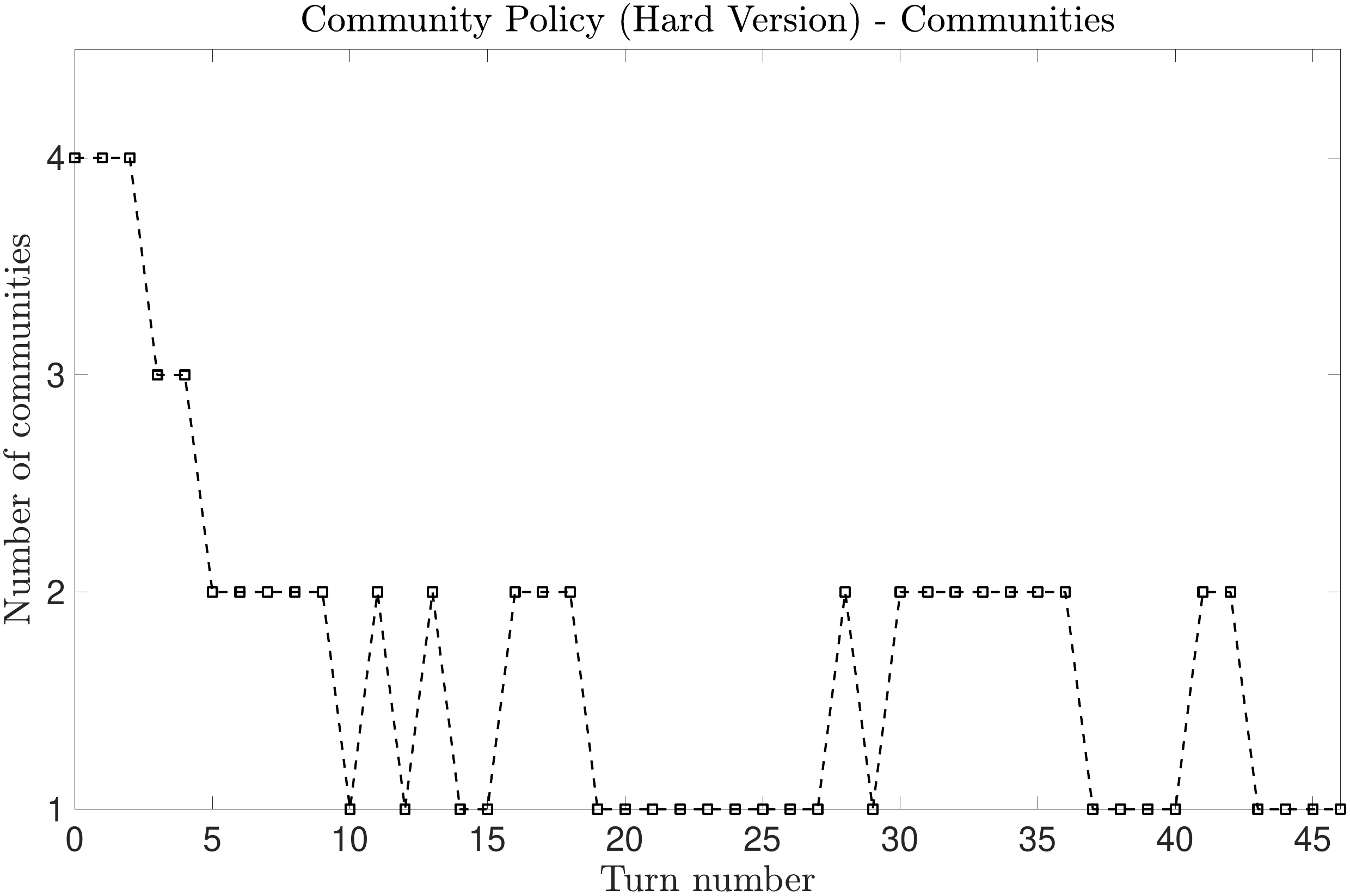}
    \caption{Comparison of the number of communities at each conversation turn when applying the Neutral policy and the Community policy in its hard version.}
    \label{fig:communities}
\end{figure}

\begin{table*}[ht]
    \label{tab:results}
    \centering
    \caption{Mean and standard deviation of the speaking time error, the number of words error, and the number of communities during the experiments performed applying respectively the five policies.}
    \begin{tabular}{c c c c c c}
        \toprule
        Policy & N  &  BH & BS & CH & CS \\
        \midrule
        Average speaking time error (s) & 40 $\pm$ 26 &  $8 \pm 5$ & $7 \pm 5$ & 19 $\pm$ 8 & 20 $\pm$ 8 \\
        Average number of words error & 66 $\pm$ 42 & 14 $\pm$ 10 & 13 $\pm$ 8 & 32 $\pm$ 10 & 33 $\pm$ 14 \\
        Average number of communities & 3.0 $\pm$ 0.5 & 1.9 $\pm$ 0.2 & 2.0 $\pm$ 0.2 & 2.0 $\pm$ 0.2 & 2.1 $\pm$ 0.4 \\    
        \bottomrule
        \end{tabular}
\end{table*}

Figure \ref{fig:communities} shows a comparison between two experiments where the robot applied the N policy and the CH policy. The x-axis represents the turn number, while the y-axis displays the number of communities formed among participants. Under the N policy, the number of communities never drops below three, indicating that each speaker often forms a separate community. Conversely, under the CH policy, the number of communities decreases from four to mostly one throughout the experiment.

Table \ref{tab:results} presents the means and standard deviations of speaking time errors, word count errors, and the number of communities across all experiments using the described policies. The error is defined as the difference at each turn between the highest and lowest values among the speakers, both for speaking times and word counts. 

To determine statistically significant differences between experiments with different policies, we used the Shapiro-Wilk test to check for normal data distribution. As some data were not normally distributed, we applied the non-parametric Mann-Whitney U test to compare the results.

We aimed to verify if the Balancing policy improved speaker participation in terms of speaking time and word count. To do this, we compared errors in speaking time and word count for the Balancing policy (BH and BS) against the baseline policy (N). Results were statistically significant with $p<0.01$, indicating a significant difference between the Neutral and Balancing policies. The average speaking time error for N was 40, compared to 8 for BH and 7 for BS, with lower standard deviations for BH and BS. No significant difference was found between BH and BS.

Next, we tested if the Community policy reduced the number of subgroups among speakers by comparing the average number of communities in CH and CS with those in N. All comparisons showed $p<0.01$, indicating significant differences when using the Community policy. We also confirmed no significant difference between CH and CS in reducing the number of communities.

Finally, we found that the Balancing and Community policies resulted in comparable outcomes regarding the average number of communities. The Mann-Whitney U test showed no significant difference among BH, BS, CH, and CS. This suggests that the Balancing policy also reduces the number of subgroups among participants. While the Balancing policy may be the most effective, selecting the best policy requires considering additional factors like the robot's preferred interaction style and user experience.


\section{Conclusion}
\label{sec:conclusion}
This paper explored the effective management of group conversations involving a social robot. Motivated by the need for systems that handle multi-party interactions, the proposed system recognizes users, engages them in dialogue, and applies policies to ensure balanced participation and foster a sense of community. This system has diverse applications in healthcare, entertainment, and education, where multi-person interaction can enhance user experience.

We implemented and tested four control policies: Balancing Hard (BH), Balancing Soft (BS), Community Hard (CH), and Community Soft (CS). These were evaluated against a baseline in 75 experiments with a humanoid robot and 300 participants grouped in sets of four. The robot moderated the conversations, applying policies to balance speaking time and word count and reduce subgroup formation. Results showed that the policies effectively managed group conversation dynamics as intended.

\bibliographystyle{plainnat}
\bibliography{main}

\end{document}